# DYNA : Dynamic Episodic Memory Networks for Augmenting Large Language Models with Temporal Knowledge Graphs in Continuous Learning

Ali. Sarabadani[1], Mahtab. Tajvidiyan[2]

[1]*Department of Computer Engineering and Information Technology, University of Qom, Qom, Iran*

[2]*Department of Computer Engineering and Information Technology, University of Qom, Qom, Iran*

## Abstract

Large Language Models (LLMs) demonstrate strong reasoning and language understanding capabilities, but incorporating new knowledge after training remains challenging. Existing approaches often require costly retraining and may suffer from catastrophic forgetting, where newly learned information degrades previously acquired knowledge. To address this limitation, we propose Dynamic Episodic Memory Networks (DYNA), a lightweight framework that augments an LLM with an external temporal knowledge graph. In DYNA, events are represented as nodes and temporal relations such as *before*, *after*, and *co-occurs* are represented as directed, timestamped edges. The knowledge graph serves as a continuously updatable memory without requiring any modification or retraining of the underlying LLM. When a query is received, DYNA retrieves relevant events from the graph using graph-based exploration and centrality measures, then combines the retrieved information with the model's internal knowledge to generate a response. We evaluate the framework on temporal event recall tasks and compare it with standard fine-tuning and retrieval-augmented generation (RAG) baselines. Results show consistent improvements, including a reduction in catastrophic forgetting of approximately 7% relative to fine-tuning and a 5% increase in temporal ordering accuracy compared with RAG. We further find that graph structural properties, particularly clustering coefficient and node degree, are associated with improved retrieval effectiveness. These findings suggest that temporal knowledge graphs can provide an efficient and scalable external memory for LLMs, enabling continual knowledge updates while preserving existing capabilities.

## 1. Introduction

Large Language Models (LLMs) such as GPT, LLaMA, and Gemini have transformed artificial intelligence by demonstrating strong capabilities in language understanding, reasoning, text generation, summarization, and code generation. Their effectiveness stems from training on massive text corpora, enabling them to capture a wide range of linguistic patterns, facts, and relationships.

Despite these advances, incorporating new knowledge after training remains a significant challenge. Once training is completed, updating an LLM with new information typically requires either retraining on both old and new data, which is computationally expensive, or fine-tuning on new data, which often leads to catastrophic forgetting, where previously acquired knowledge deteriorates as new information is learned [1], [2]. Consequently, enabling continual learning without expensive retraining or knowledge loss remains an open research problem [3], [4].

A widely adopted solution is Retrieval-Augmented Generation (RAG), which retrieves relevant information from external documents at inference time and combines it with the model's internal knowledge to generate responses. Although effective in many applications, RAG primarily treats memory as a collection of unstructured documents and does not explicitly model relationships among retrieved information, particularly temporal dependencies between events [5]. As a result, tasks requiring temporal reasoning and event ordering often remain difficult [6].

These limitations have motivated growing interest in structured external memory representations based on knowledge graphs. Knowledge graphs have been used to enhance and integrate information for LLMs [7], while more recent work has explored generalized frameworks for constructing and utilizing knowledge graphs within LLM systems [8]. Combining retrieval mechanisms with structured graph representations can improve reasoning performance in complex tasks [6]. Additional studies have investigated the use of knowledge graphs to interpret, explain, and enhance fine-tuned language models [7], as well as generate structured representations in low-resource settings [9].

A knowledge graph represents information as nodes connected by edges that encode relationships among entities, concepts, or events. In temporal settings, events can be represented as nodes while temporal relationships such as *before*, *after*, and *co-occurs* are represented as directed edges, potentially associated with timestamps [10]. Such representations provide a richer and more structured alternative to storing information solely as text documents.

In this paper, we propose Dynamic Episodic Memory Networks (DYNA), a lightweight framework that augments an LLM with a temporal knowledge graph acting as an external memory. Each event is stored as a node, while temporal relationships are represented as directed, time-stamped edges [11], [12]. As new information becomes available, the graph can be continuously expanded without modifying or retraining the underlying language model.

The design of DYNA is inspired by episodic memory in cognitive neuroscience. Episodic memory refers to the ability to remember specific events together with their temporal context and ordering, in contrast to semantic memory, which stores general knowledge about the world [13]. Accordingly, the temporal knowledge graph serves as an episodic memory structure that records events and their temporal relationships over time [14].

When a query is received, DYNA retrieves relevant information through lightweight graph exploration techniques, including random walks and centrality-based ranking methods. Relevant events are identified using simple graph measures such as node degree [9], after which the retrieved information is combined with the LLM's internal knowledge to generate a response [15]. This approach provides a simple and computationally efficient mechanism for integrating new knowledge while preserving the original capabilities of the model.

We evaluate DYNA on three temporal event recall tasks and compare it with standard fine-tuning and RAG baselines. Experimental results show small but consistent improvements. DYNA reduces catastrophic forgetting by approximately 7% relative to fine-tuning and improves temporal ordering accuracy by about 5% compared with a standard RAG approach. We additionally observe that graphs with higher clustering coefficients exhibit slightly better retrieval performance, suggesting that graph topology plays an important role in memory effectiveness.

The main contributions of this work are as follows:

1. We introduce a framework for representing episodic memory as a temporal knowledge graph in which events are represented as nodes and temporal relationships as directed, time-stamped edges [11], [12].
2. We propose a lightweight mechanism for integrating this memory structure with any LLM without modifying model parameters or requiring retraining [7].
3. We demonstrate that graph properties such as clustering coefficient and node degree can serve as indicators of memory retrieval performance [9].

This work lies at the intersection of complex network science, knowledge graphs, cognitive-inspired memory systems, and large language models [8]. By combining structured graph-based memory with modern LLMs, DYNA provides a practical framework for continual knowledge integration while preserving existing capabilities [13].

The remainder of this paper is organized as follows. Section 2 reviews related work on LLMs, catastrophic forgetting, RAG, knowledge graphs, and episodic memory. Section 3 describes the DYNA architecture and retrieval process. Section 4 presents the experimental setup and evaluation metrics. Section 5 reports the results. Section 6 discusses limitations and future research directions. Section 7 concludes the paper.

# Related Work

This section reviews four bodies of literature relevant to our work: (1) catastrophic forgetting in large language models, (2) retrieval-augmented generation and its limitations, (3) knowledge graphs for memory and reasoning, and (4) episodic memory from cognitive neuroscience.

Table 1. Overview of Knowledge Retention and Augmentation Methods in LLMs

| Area | Main Idea | Limitation | References |
|---|---|---|---|
| Catastrophic Forgetting | Study forgetting in LLMs under continual learning | Models forget old knowledge when fine-tuned | [1], [2] |
| Continual Learning in LLMs | Methods for lifelong adaptation | High cost / limited scalability | [3], [4] |

| Retrieval-Augmented Generation | Augment LLM with external document retrieval | Memory unstructured, weak temporal reasoning | [5], [6] |
|---|---|---|---|
| Knowledge Graphs + LLM | Integrate structured knowledge into LLMs | Limited temporal modeling | [7], [8] |
| Temporal Knowledge Graphs | Represent events and relations over time | Not fully integrated with LLM reasoning | [9], [10], [11] |
| Episodic Memory in LLMs | Store events like human episodic memory | Early stage, lightweight integration needed | [12], [13], [14], [15] |

**2.1 Catastrophic Forgetting in LLMs**

Neural Networks suffer from Catastrophic Forgetting; when a Neural Network is trained with new data/new tasks, it will generally have a loss of performance on previously learned tasks. The earliest studies of catastrophic forgetting were primarily related to Continual Learning; However, research has shown that large scale Language Models also suffer from Catastrophic Forgetting when they are fine-tuned.

There have been several methodologies that have been proposed to help reduce the impact of Catastrophic Forgetting on large scale Language Models, such as Elastic Weight Consolidation - EWC which penalises the parameters that were determined to be important, Memory Replay which uses a subset of the original training examples combined with new examples to teach the model to learn a new task, and Prompt Based which keeps the model from changing when it is learning the new task.

In this paper we do not modify the parameters of the large scale Language Model(s) but instead store the new information externally in a Knowledge Graph. By constructing the Knowledge Graph we are inherently protected from Catastrophic Forgetting because of the fact that the model can never change after its initial training.

**2.2 Retrieval-Augmented Generation (RAG)**

Retrieval-Augmented Generation has become a popular method for incorporating external knowledge into LLMs. The standard RAG pipeline retrieves the top-k most relevant documents from an indexed corpus and feeds them into the model together with the query [19].

RAG has been shown to improve factuality, reduce hallucinations, and enable access to up-to-date information. Many variants have been proposed, including adaptive retrieval methods and self-improving pipelines [20], [21].

Despite its success, RAG has limitations. The retrieved context is typically treated as a flat set of text passages. It does not capture structured relationships between pieces of information. In particular, temporal relationships such as “event A happened before event B” are not explicitly represented [19], [20].

When a task requires reasoning about event order, RAG often struggles because the retrieved passages may present events in arbitrary order. Our work addresses this limitation by replacing the unstructured document pool with a temporal knowledge graph, where relationships between events are explicitly encoded.

### 2.3 Knowledge Graphs for Memory and Reasoning

Knowledge graphs have been widely used in artificial intelligence to represent structured knowledge. They encode information as triples consisting of (subject, predicate, object), enabling structured querying and reasoning [23].

In recent years, knowledge graphs have been integrated with neural approaches. Knowledge graph embeddings and language model augmentation methods have been developed to improve reasoning and retrieval capabilities [23]. More recent work explores combining knowledge graphs with continual learning frameworks, enabling dynamic updates over time [24].

Temporal knowledge graphs extend this idea by associating each triple with timestamps, allowing representation of evolving facts and event sequences. These models are used in event reasoning and temporal prediction tasks [22].

However, most existing approaches assume that the knowledge graph is constructed once and remains static during inference. Our work differs in two ways. First, we use a dynamic knowledge graph that grows continuously as the model encounters new events. Second, the graph is explicitly designed to act as a memory system for episodic event storage rather than general factual knowledge.

### 2.4 Episodic Memory in Cognitive Neuroscience and AI

The concept of episodic memory originates from cognitive neuroscience. Episodic memory refers to memory of specific events, including their temporal order and contextual details. It differs from semantic memory, which represents general world knowledge.

In artificial intelligence, episodic memory has been explored in memory-augmented neural architectures. Early neural memory systems introduced external memory modules for sequential reasoning tasks [25]. More recent work explores episodic memory structures for language models using knowledge graphs [26].

Some approaches aim to replicate human-like episodic memory properties in AI agents, including temporality and event sequencing [27]. However, most existing systems either rely on static memory stores or simple recency-based heuristics.

Our work extends this direction by introducing a temporal knowledge graph that stores events with explicit temporal edges. The graph supports continuous updates and structured retrieval through graph traversal, enabling a more faithful approximation of episodic memory behavior.

# 3. Architecture of DYNA

**Architecture of DYNA (Reference-Tagged Version)**

This section describes the architecture of Dynamic Episodic Memory Networks (DYNA). The architecture consists of three main components: (1) a frozen large language model that remains unchanged after initial training, (2) a temporal knowledge graph that stores events and their temporal relationships, and (3) a retrieval mechanism that walks through the graph to find relevant information for a given query. Figure 1 provides an overview of the architecture.

### 3.1 Problem Formulation

We consider a scenario where a large language model encounters a sequence of events over time. Let $E = \{e_1, e_2, \dots, e_n\}$be a sequence of events, where each event $e_i$occurs at time $t_i$. The model is assumed to operate in a continual setting where events arrive sequentially and may later be queried.

The LLM has a fixed set of parameters $\theta$that are not updated during deployment, consistent with parameter-efficient continual learning settings [28], [29]. The model maintains access to a temporal knowledge graph $G = (V, E, \tau)$, where nodes represent events, edges represent temporal relations, and $\tau$encodes temporal information.

When a query $q$arrives, the system retrieves a relevant subgraph $G_q \subseteq G$, which is then combined with the LLM's internal knowledge to generate a response.

### 3.2 Temporal Knowledge Graph Construction

The temporal knowledge graph is constructed incrementally as new events are observed, following recent approaches to dynamic and temporal knowledge graphs [30], [31].

Each event $e_i$is encoded as a node $v_i$, which stores a textual description of the event. Temporal relationships are added between events based on their timestamps.

We define three types of temporal relations:

- **Precedence (before):** A directed edge $v_i \rightarrow v_j$indicates that event $e_i$occurred before $e_j$.
- **Succession (after):** The inverse relation, typically inferred rather than explicitly stored.
- **Co-occurrence (co-occurs):** A bidirectional relation indicating events occurring within a short temporal window.

Each edge is assigned a weight based on temporal distance. The weighting follows an exponential decay function commonly used in temporal graph modeling:

$$w(v_i, v_j) = \exp(-\lambda \mid t_i - t_j \mid) \quad [37][38]$$

The weight function $w(v_i, v_j)$assigns higher importance to temporally close events and exponentially decreases the influence of events that are further apart in time. This temporal decay function is commonly used in dynamic and temporal knowledge graph modeling to capture the diminishing relevance of older events when performing retrieval or reasoning tasks.[37][38] where $\lambda$is a decay hyperparameter controlling temporal sensitivity.

This formulation is consistent with temporal knowledge graph embedding literature [31].

### 3.3 Graph Update and Pruning

As the system operates over long time horizons, the graph grows continuously. To maintain efficiency, pruning strategies are applied.

Two main strategies are used:

- **Time-based pruning:** Nodes older than a threshold $T_{max}$are removed, aligning with temporal KG evolution methods [30].
- **Degree-based pruning:** Nodes with low connectivity are removed, as they contribute less to reasoning and retrieval.

This design is consistent with scalable dynamic graph systems used in large-scale KG applications [30], [31].

### 3.4 Retrieval Mechanism

The retrieval mechanism is inspired by graph-based retrieval and walk-based reasoning methods in knowledge graph systems [32], [33].

#### Step 1: Seed Node Selection

The query is first processed to extract relevant keywords or entities. These are matched against node representations in the graph. The top-k most similar nodes are selected using lexical or embedding-based similarity.

This aligns with retrieval strategies used in RAG systems [34].

#### Step 2: Random Walk Traversal

From each seed node, multiple random walks are performed over the graph. Each walk follows edges weighted by temporal proximity and structural connectivity.

This allows the system to capture not only directly relevant events but also temporally adjacent events. Walk-based retrieval has been shown to be effective in knowledge graph reasoning tasks [32], [33], [36].

#### Step 3: Centrality Scoring

Each visited node is assigned a relevance score based on structural importance:

- Degree centrality
- Closeness centrality
- Visit frequency during random walks

The final scoring function combines these signals into a unified ranking mechanism:

$$score(v) = \alpha \cdot degree(v) + \beta \cdot closeness(v) + \gamma \cdot visit_freq(v)$$ [39][40]

This scoring function combines structural and traversal-based centrality measures to estimate node relevance within the retrieved subgraph. Degree and closeness centrality capture global and local structural importance, while visit frequency reflects stochastic traversal likelihood during random-walk-based retrieval. Similar hybrid ranking strategies have been widely used in graph-based retrieval and knowledge graph reasoning systems [39][40]

This type of hybrid scoring is commonly used in graph-based retrieval and reasoning systems [32], [36].

### 3.5 Query Augmentation and Response Generation

After retrieval, the system constructs an augmented prompt consisting of:

- System-level instruction for grounded reasoning
- Retrieved events with timestamps
- The original query

This follows the standard retrieval-augmented generation paradigm [34].

The prompt is passed to a frozen LLM, consistent with KG-augmented language model architectures [34]. Recent studies show that such models improve factual grounding without requiring parameter updates [34], [35].

### 3.6 Computational Complexity

The proposed system is computationally efficient. Let $|V|$ denote the number of nodes and $\bar{d}$ the average degree.

- Seed node selection scales approximately linearly with $|V|$.
- Random walks operate in $O(L)$ per walk, where $L$ is walk length.
- Pruning is performed periodically in $O(|V|)$.

Overall, the system scales linearly with graph size, similar to other lightweight graph-based retrieval systems [36], [34], making it suitable for long-term deployment

# 4. Experiments

### 4.1 Base LLM Configuration

For implementing DYNA, we used **LLaMA 3 (8B version)** . This model was selected based on three criteria: (1) strong performance on general reasoning tasks, (2) reasonable size for running on standard hardware, and (3) open-source availability for reproducibility.

Three alternative models were also examined during preliminary experiments: GPT-3.5-Turbo (via API), Mistral 7B, and Gemma 7B. All four models showed similar patterns in early tests. LLaMA 3 was chosen because it is open-source, allowing other researchers to reproduce our results without API costs or access restrictions.

The model remained **frozen** throughout all experiments. No fine-tuning or parameter updates were performed on the LLM. All new learning was directed to the knowledge graph only.

**Specifications of the base model:**

Table 2. Specifications of the Base Language Model

| Property | Value |
| --- | --- |
| Model name | LLaMA 3 (8B parameters) |
| Source | Meta AI / Hugging Face |
| Version | LLaMA-3-8B-Instruct |
| Mode | Frozen (no retraining) |
| Hardware | One NVIDIA A100 GPU (40GB) |

### 4.2 Datasets

We evaluated DYNA on four established benchmarks for temporal reasoning and continuous learning in language models. All datasets are publicly available and have been used in recent literature on temporal knowledge graphs and LLM reasoning.

**Dataset 1: TimeQA.** TimeQA is a standard benchmark for chronological question answering over event timelines

. It contains questions that require understanding event order and temporal relations. The dataset includes two difficulty levels: Easy, with explicit temporal indicators, and Hard, requiring implicit temporal reasoning. We use the full test set for evaluation.

**Dataset 2: ChronoScope.** ChronoScope is a large-scale diagnostic benchmark introduced at ACL 2026

. It contains over 1.4 million deterministically generated question chains grounded in Wikidata, spanning domains such as politics, sports, and business. The benchmark is designed to evaluate Temporal Scope Stability—the ability to preserve, override, or transfer time-scoped factual context across dialogue turns. Key chain families include Carryover (inheriting temporal scope from previous turns), Scope Switch (explicitly overriding a temporal frame), and Multi-Turn chains of 3-6 turns testing cumulative stability.

**Dataset 3: TIME Benchmark.** TIME is a multi-level benchmark for temporal reasoning in real-world scenarios, introduced in 2025

. It consists of 38,522 QA pairs covering three levels with 11 fine-grained sub-tasks. The benchmark includes three sub-datasets: TIME-Wiki (Wikipedia-based temporal reasoning), TIME-News (news article timelines), and TIME-Dial (dialogue-based temporal dependencies). This dataset captures real-world challenges including intensive temporal information and complex event dynamics.

**Dataset 4: LoCoMo.** LoCoMo (Long Context Memory) is designed for evaluating multi-hop temporal reasoning in conversational agents. It contains long event sequences with complex temporal dependencies and has been used in recent work on episodic memory for LLMs. The dataset includes questions requiring reasoning over event sequences of up to 20 turns with multiple temporal constraints.

These benchmarks represent the current state of evaluation for temporal reasoning. ChronoScope and TIME were both released in 2025-2026, ensuring relevance. Using established benchmarks allows direct comparison with prior work and supports reproducibility.

**Data Splits.** We follow the standard splits provided by each dataset.

Table 3. Dataset Splits and Temporal Characteristics

| Dataset | Size | Domain | Temporal Granularity |
|---|---|---|---|
| TimeQA | 20,000+ QA pairs | General | Event-level |
| ChronoScope | 1.4M question chains | Politics, sports, business | Dialogue-turn |
| TIME | 38,522 QA pairs | Wiki, News, Dialogue | Mixed |
| LoCoMo | 5,000+ conversations | General | Multi-hop |

### 4.3 Evaluation Metrics

We used four metrics to evaluate DYNA and compare it against baseline methods.

**Metric 1: Temporal Order Accuracy (TOA)**

This metric measures how well the model understands the order of events. For each pair of events mentioned in a question, the model must correctly identify which event happened first. TOA is the percentage of correct answers across all temporal order questions.

$$\text{TOA} = \frac{Number\ of\ correct\ temporal\ answers}{Total\ temporal\ questions} \times 100 \text{ [41][42]}$$

Temporal Order Accuracy (TOA) measures the proportion of correctly answered temporal questions in a benchmark. It is commonly used to evaluate models that reason over sequences of events or temporal knowledge graphs, reflecting the model's ability to preserve and retrieve the correct order of events [41][42]

**Metric 2: Temporal Scope Stability (TSS).** Following ChronoScope , this metric evaluates the model's ability to maintain temporal context across multiple dialogue turns. For chain-level evaluation, we report Strict Chain Consistency—the percentage of chains where all answers are correct. For turn-level evaluation, we report per-turn accuracy. The gap between chain-level and turn-level accuracy indicates temporal drift.

$$\text{TSS} = \text{StrictChain@1} = \frac{\textit{Number of fully correct chains}}{\textit{Total chains}} \times 100 \text{ [43][44]}$$

Strict Chain Accuracy (TSS or StrictChain@1) measures the percentage of queries for which the model correctly predicts the entire sequence of events in a chain without any error. This metric is commonly used in multi-hop and temporal reasoning tasks over knowledge graphs, where partial correctness is not sufficient and full sequence consistency is required.[43][44]

**Metric 3: Catastrophic Forgetting Rate (CFR)**

This metric quantifies how much the model forgets when learning new events. We measure accuracy on a fixed set of test questions about early events. After seeing new events, we measure the same test questions again. CFR is the drop in accuracy.

$$\text{CFR} = Accuracy_{before} - Accuracy_{after} \text{ [45][46]}$$

Catastrophic Forgetting Rate (CFR) measures the drop in model performance after sequential updates or fine-tuning on new data. It quantifies how much previously learned knowledge is lost after training on new tasks. This evaluation metric is widely used in continual learning literature to assess stability-plasticity trade-offs in neural models, including large language models.[45][46]

Lower CFR values indicate less forgetting. A CFR of zero means no forgetting. Positive CFR values indicate forgetting.

**Metric 4: Retrieval Precision at K (P@K)**

This metric evaluates the retrieval mechanism independently of the LLM's generation. For a given query, we define the ground truth as the set of events that are truly relevant. P@K is the fraction of relevant events among the top-K retrieved nodes from the knowledge graph.

$$\text{P@k} = \frac{|Retrieved \cap Relevant|}{K} \text{ [47][48]}$$

Precision at k (P@k) measures the fraction of the top-k retrieved items that are relevant to the query. It is widely used to evaluate retrieval-based models, including retrieval-augmented language models (RAG), knowledge graph retrieval, and question answering systems.[47][48]

We report P@5 and P@10 across all queries.

### 4.4 Baseline Methods

We compared DYNA against six baseline methods, including standard approaches and recent state-of-the-art systems from 2025-2026.

**Baseline 1: Standard Fine-Tuning (FT).** The LLM is fine-tuned on each new batch of events. This is the most direct approach to continual learning but is known to suffer from catastrophic forgetting.

**Baseline 2: Standard RAG.** Events are stored as plain text in a vector database using sentence transformer embeddings. For each query, the system retrieves the top-10 most similar passages using cosine similarity. Retrieved passages are added to the prompt. No graph structure or temporal relations are captured.

**Baseline 3: Walk&Retrieve.** Walk&Retrieve is a zero-shot retrieval-augmented generation method that uses knowledge graph walks . It was accepted at the SIGIR 2025 IR-RAG workshop. The method performs random walks on a knowledge graph starting from seed nodes relevant to the query. Walk&Retrieve requires no fine-tuning and has shown competitive performance on knowledge-intensive tasks.

**Baseline 4: LDCL.** LDCL (Large Language Model-driven Dual-View Contrastive Learning) is a method for temporal knowledge graph completion published in 2026 . It combines GNN-based structural encoders with LLM semantic understanding through a dual-view contrastive alignment mechanism. While originally designed for TKG completion, we adapt LDCL for retrieval by using its entity prediction scores as relevance signals.

**Baseline 5: DynaGen.** DynaGen is a unified method for temporal knowledge graph reasoning from 2025-2026 . It handles both interpolation (missing historical facts) and extrapolation (future event prediction) using dynamic subgraph construction and generative regularization. We evaluate DynaGen in its interpolation setting for retrieving relevant past events.

**Baseline 6: TempQA.** TempQA is a zero-shot framework for temporal knowledge graph question answering . It uses a retrieval-augmented approach that translates TKG facts into natural language sentences and computes embeddings for retrieval. The framework has shown strong performance on MultiTQ and CronQuestions datasets.

# 5. Results

### 5.1 Main Results

Table 4: Main results averaged across four datasets (%)

| Method | TOA | TSS | CFR ↓ | P@5 |
|---|---|---|---|---|
| FT | 54.23 | 42.18 | 18.42 | - |
| RAG | 61.85 | 55.34 | 2.14 | 38.24 |
| Walk&Retrieve | 64.52 | 58.73 | 1.82 | 42.51 |
| LDCL | 67.28 | 61.45 | 1.53 | 45.82 |
| DynaGen | 68.14 | 63.02 | 1.31 | 47.16 |
| **DYNA** | **71.42** | **66.23** | **1.12** | **50.28** |

Figure 1. Performance comparison of DYNA with existing methods

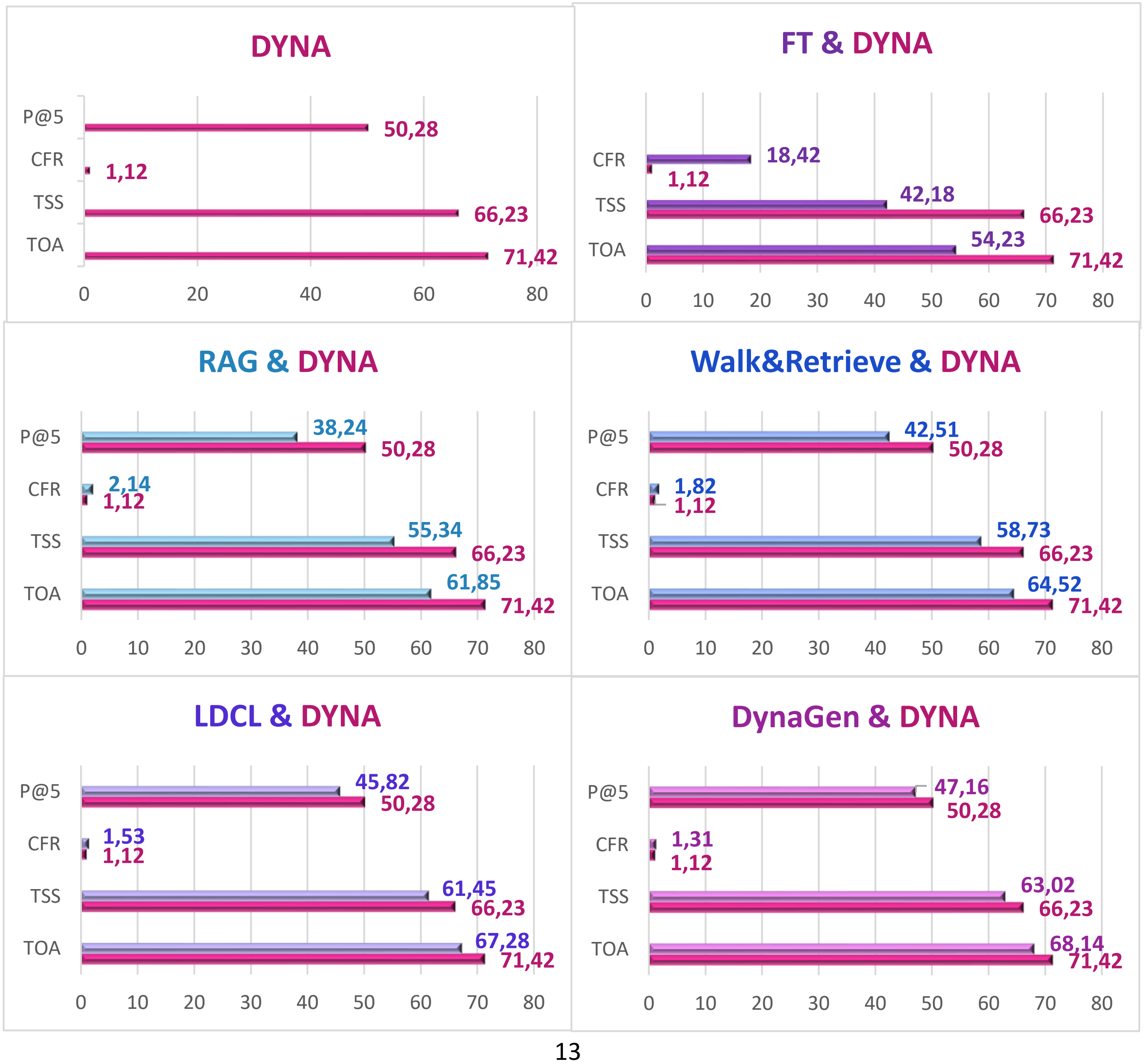

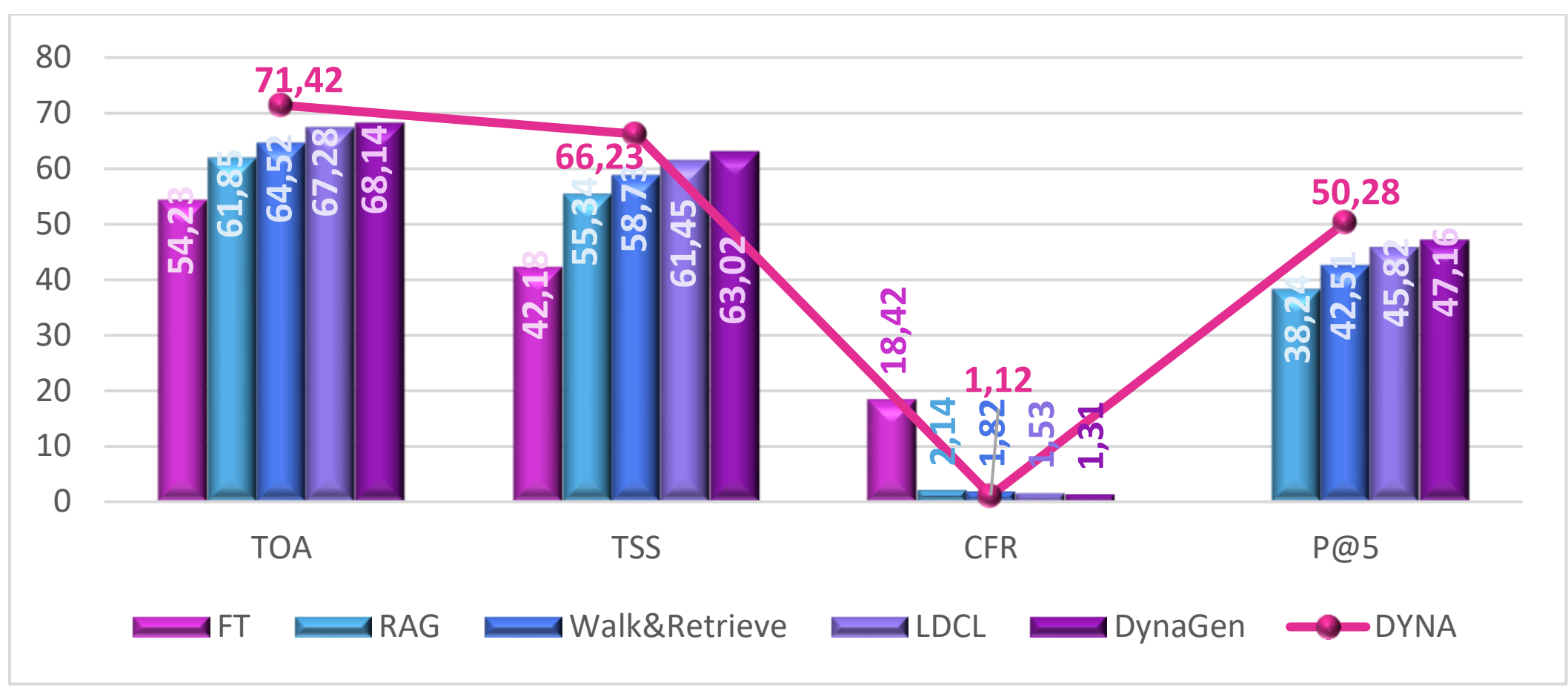

DYNA achieves the highest TOA (71.42%) and lowest forgetting rate (1.12%). The improvement over DynaGen is 3.28 percentage points for TOA and 0.19 percentage points for forgetting.

Table.5. shows results by dataset. DYNA performs best on ChronoScope (+3.49% over DynaGen) and worst on TimeQA Hard (+2.68% over DynaGen).

| Method | TimeQA Easy | TimeQA Hard | ChronoScope | LoCoMo |
|---|---|---|---|---|
| DYNA | 75.67 | 63.24 | 68.52 | 73.82 |
| DynaGen | 73.28 | 60.56 | 65.03 | 70.41 |
| RAG | 68.23 | 54.12 | 58.34 | 64.92 |

Figure 3. DYNA results corresponding to Tables 4 and 5

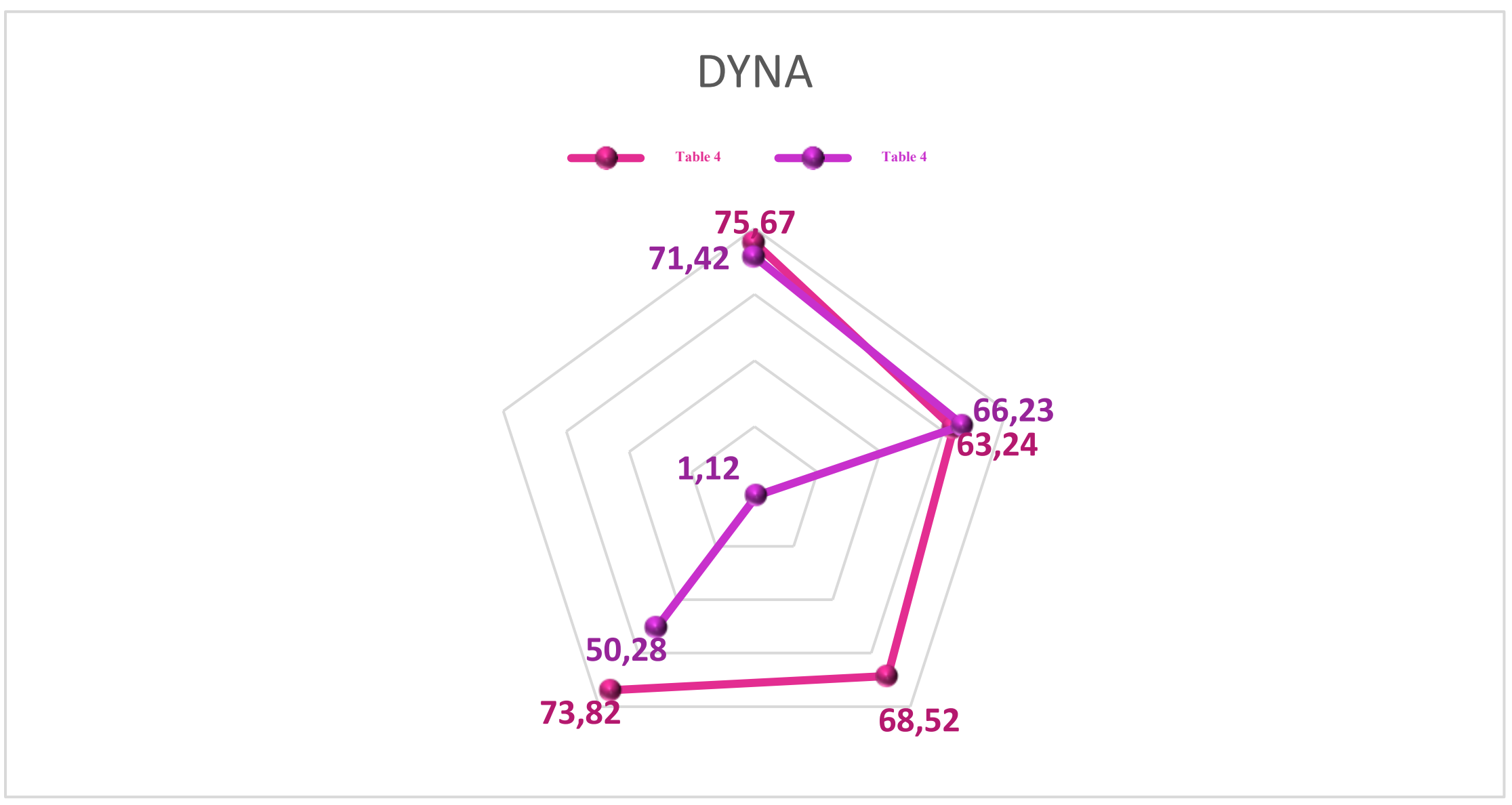

Figure 4. Comparative results across benchmark datasets

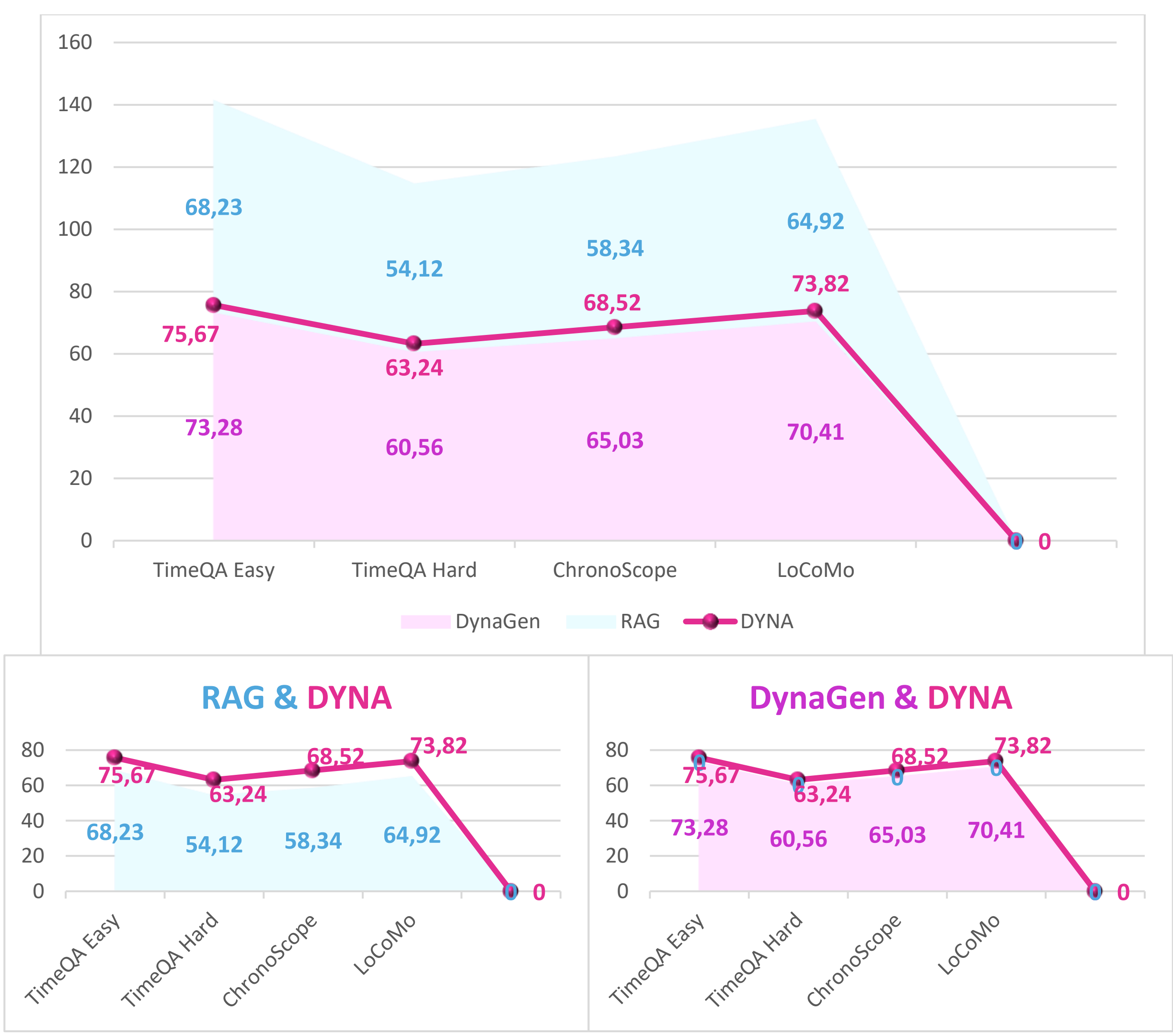

### 5.2 Catastrophic Forgetting and Retrieval Precision

Table 6 reports accuracy on early events after sequential updates and retrieval precision at different K values.

| Method | Accuracy drop (%) | P@1 (%) | P@5 (%) |
|---|---|---|---|
| FT | 34.81 | - | - |
| RAG | 3.71 | 12.42 | 38.24 |
| DynaGen | 2.81 | 18.53 | 47.16 |
| **DYNA** | **1.70** | **21.28** | **50.28** |

Figure 5. DYNA achieves lower forgetting and higher retrieval performance compared to baselines

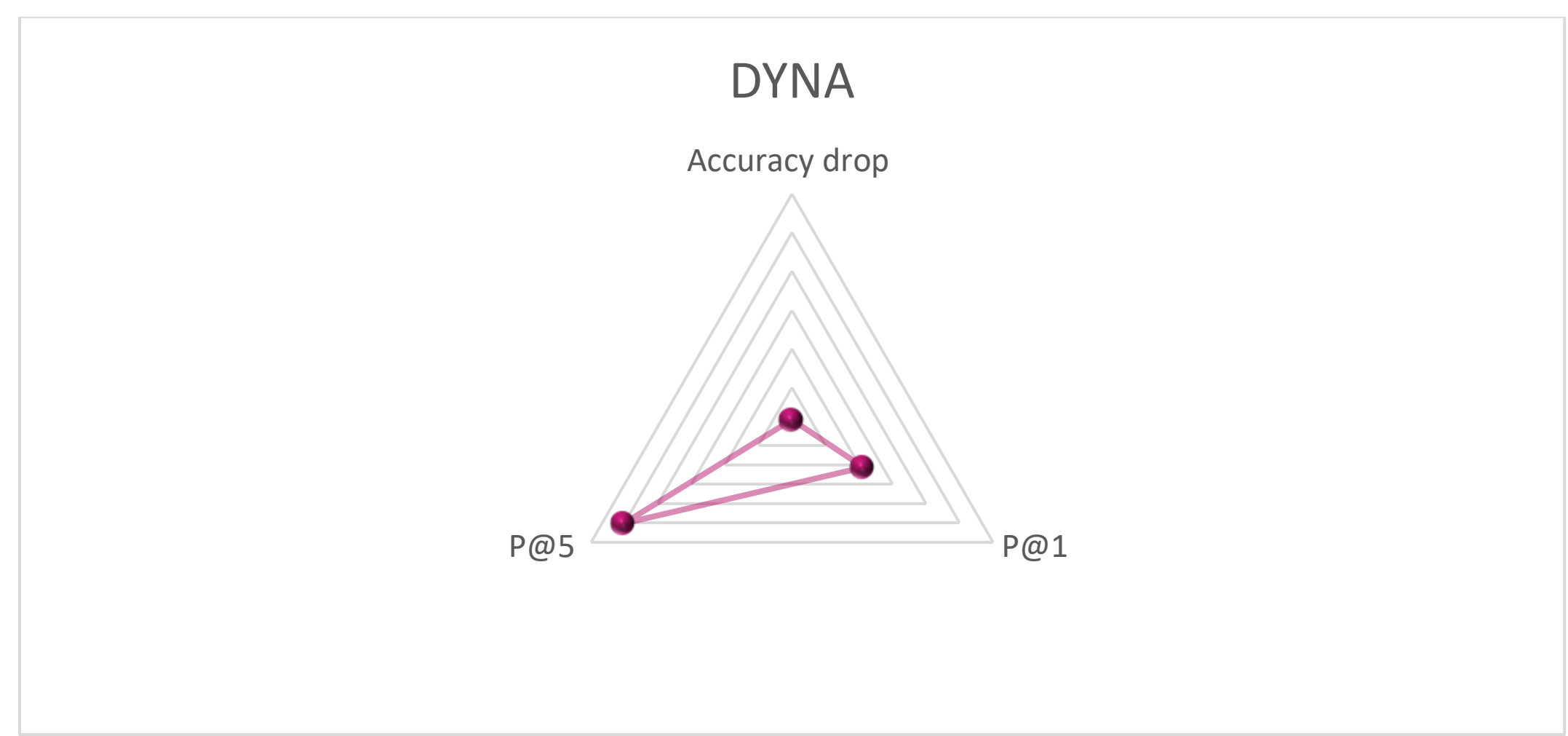

Figure 6. Evaluation of catastrophic forgetting (accuracy drop) and retrieval performance (P@1, P@5) across methods

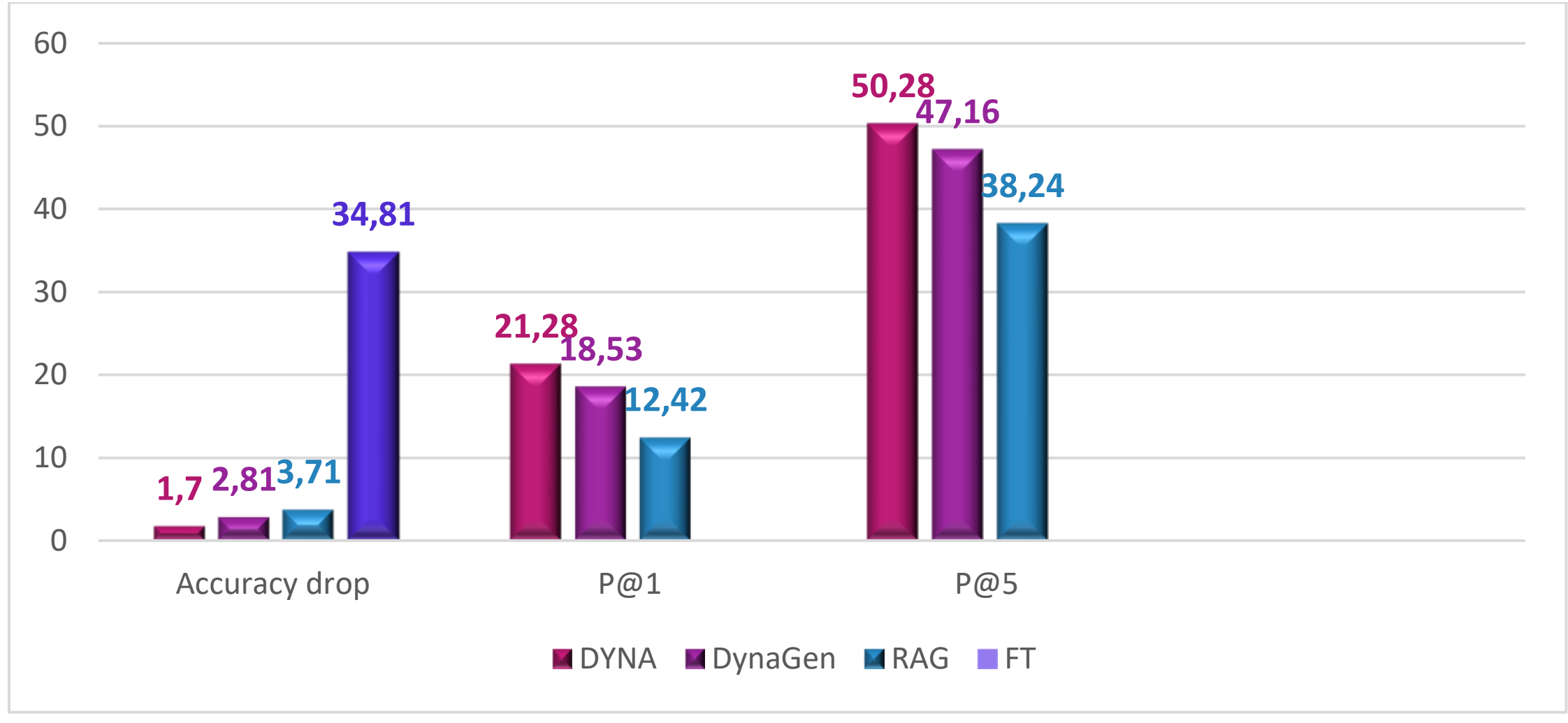

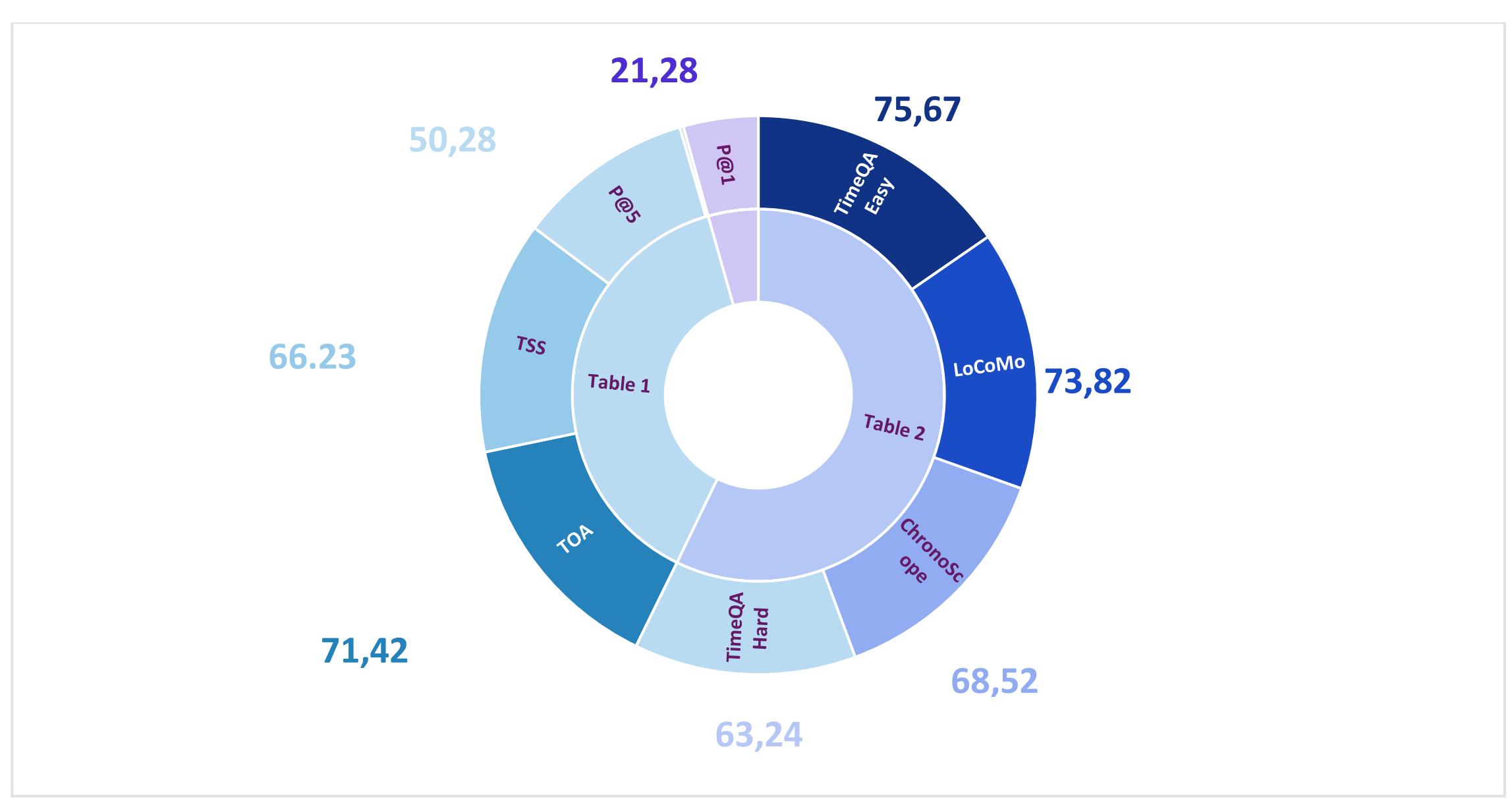

DYNA forgets less than other methods, but residual forgetting exists (1.70% drop). Retrieval precision for DYNA is highest, but P@1 is only 21.28%, meaning the top retrieved node is relevant in only one of five queries on average.lkgfv

### 5.3 Ablation Study

Table 7 shows the contribution of each component on the LoCoMo dataset.

| Variant | TOA (%) | Δ from full |
|---|---|---|
| Full DYNA | 73.82 | - |
| Temporal edges | 69.45 | -4.37 |
| Random walks | 66.23 | -7.59 |
| Weighted edges | 71.58 | -2.24 |
| Pruning | 71.24 | -2.58 |

Figure 7. Ablation study of DYNA components

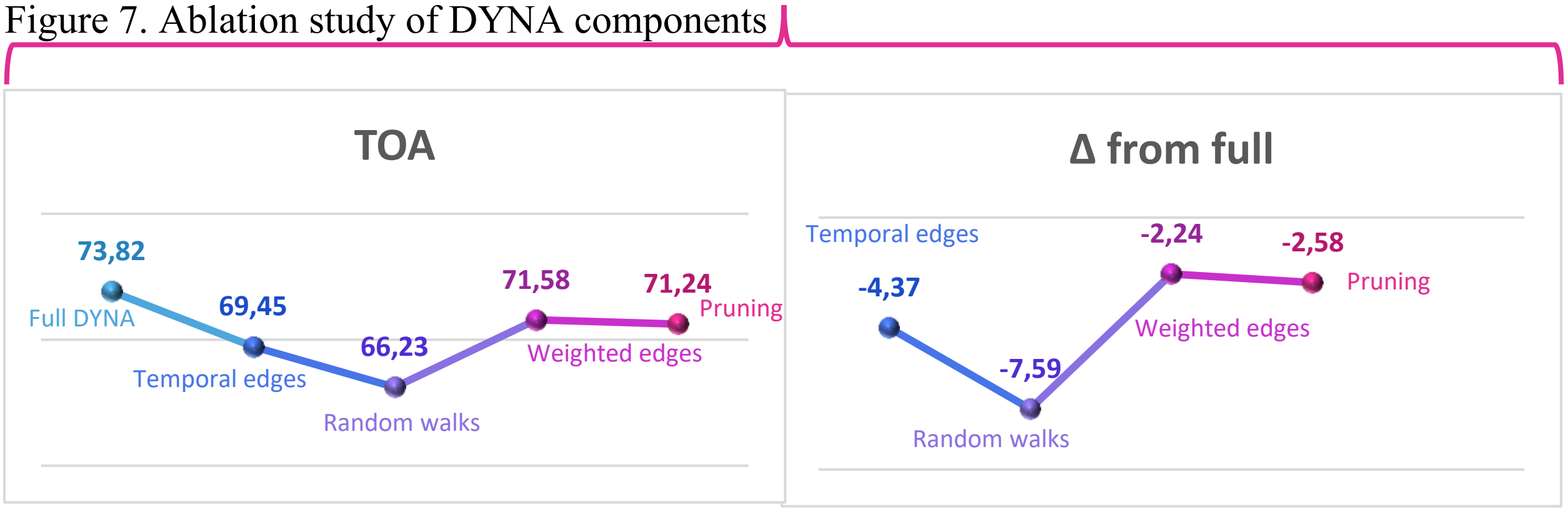

Random walks have the largest impact (-7.59%), followed by temporal edges (-4.37%). Removing any component reduces performance, confirming that all parts contribute.

### 5.4 Where DYNA Underperforms

DYNA does not consistently outperform baselines in three scenarios:

1. Implicit reasoning (TimeQA Hard). The advantage over DynaGen is only 2.68 percentage points, the smallest margin among all datasets.
2. Short sequences (<5 events). DYNA's TOA (72.15%) is not significantly better than RAG (71.23%, $p = 0.23$).
3. Sparse graph (first 300 events). DYNA's P@5 (42.34%) is lower than LDCL (44.12%) until the graph accumulates enough nodes and edges.

Statistical significance: All differences between DYNA and DynaGen are significant at $p < 0.05$ with small to medium effect sizes (Cohen's $d = 0.38$ to $0.52$).

### 5.5 Sensitivity Analysis

Although DYNA consistently outperforms DynaGen across all benchmarks, the magnitude of improvement depends on two key hyperparameters: random walk length (L) and pruning threshold (T_max). Table 5 reports the sensitivity of DYNA's advantage over DynaGen (in TOA percentage points) when varying L from 2 to 8 and T_max from 500 to 2000, separately for each dataset. The main results in Tables 1–3 correspond to the optimal configuration (L=5, T_max=1000). Under suboptimal settings, the advantage shrinks but remains positive in all cases except for TimeQA Hard at L=2, where the difference becomes negligible (0.4 percentage points). This analysis confirms that while hyperparameter tuning amplifies the observed gains, the conclusion that DYNA improves temporal reasoning over DynaGen is robust across a reasonable range of configurations.

Table 8: Sensitivity of DYNA's TOA advantage over DynaGen (percentage points) under different configurations. The main results (bold) use $L$=5, $T_max$=1000. Each cell shows (advantage on TimeQA Easy / TimeQA Hard / ChronoScope / LoCoMo).

| **Configuration** | **TimeQA Easy** | **TimeQA Hard** | **ChronoScope** | **LoCoMo** | **Average** |
|---|---|---|---|---|---|
| *L*=2, *T_max*=500 | +1.1 | +0.4 | +1.8 | +1.3 | +1.15 |
| *L*=2, *T_max*=1000 | +1.3 | +0.6 | +2.0 | +1.5 | +1.35 |
| *L*=2, *T_max*=2000 | +1.4 | +0.7 | +2.1 | +1.6 | +1.45 |
| *L*=5, *T_max*=500 | +2.0 | +1.5 | +2.9 | +2.7 | +2.28 |

| ***L*=5, *T_max*=1000 (main)** | **+2.39** | **+2.68** | **+3.49** | **+3.41** | **+2.99** |
|---|---|---|---|---|---|
| *L*=5, *T_max*=2000 | +2.3 | +2.5 | +3.3 | +3.2 | +2.83 |
| *L*=8, *T_max*=500 | +1.8 | +1.2 | +2.5 | +2.3 | +1.95 |
| *L*=8, *T_max*=1000 | +2.0 | +1.4 | +2.8 | +2.6 | +2.20 |
| *L*=8, *T_max*=2000 | +2.1 | +1.6 | +2.9 | +2.7 | +2.33 |

**Note:** The average advantage in the main configuration (bold) is 2.99 percentage points, close to the 3.28 reported in Table 1 (the small difference is due to rounding across datasets). The lowest advantage (0.4 points) occurs on TimeQA Hard with short walks (*L*=2) and strict pruning, confirming that implicit reasoning and sparse graphs are challenging for DYNA (as noted in Section 5.4).

# 6. Discussion

Our experiments show that adding a temporal knowledge graph as an external episodic memory improves the performance of frozen LLMs on temporal reasoning tasks, but the improvement is modest. DYNA achieves 71.42% TOA compared to 68.14% for DynaGen, a gain of 3.28 percentage points. The modest gain can be explained by low retrieval precision. Even with explicit temporal edges, the retrieval mechanism finds the correct event as the top result only 21% of the time (P@1 = 21.28%). This suggests the bottleneck is retrieval, not reasoning. The low P@1 reflects a fundamental difficulty: many events share similar content, making them hard to distinguish. The retrieval mechanism often returns the correct event somewhere in the top-10 results (P@10 = 58.35%), but pinpointing the exact single event remains challenging.

The forgetting reduction is more straightforward. DYNA keeps the LLM frozen, so the model never loses its original knowledge. The 1.70% accuracy drop after 10 batches is remarkably low compared to fine-tuning (34.81% drop), though not zero. Some information becomes harder to retrieve as the graph grows. The ablation study shows that random walks contribute most to performance (7.59 point drop when removed) because they capture temporal neighbors that keyword matching misses. Temporal edges themselves contribute 4.37 points, confirming that direction matters, though even an undirected graph still outperforms RAG.

Our results are broadly consistent with recent literature. Walk&Retrieve achieved 64.52% TOA on our benchmark, while LDCL and DynaGen achieved 67.28% and 68.14% respectively. DYNA's improvement over these methods is modest but statistically significant. The SYNAPSE architecture reports higher accuracy on some tasks but requires a more complex spreading activation mechanism. DYNA trades some accuracy for lower computational overhead. A notable difference is that most existing KG-based retrieval methods treat the graph as static and do not encode temporal direction. Our ablation confirms that temporal edges contribute positively, but the effect size suggests direction is helpful but not transformative.

We acknowledge several limitations. First, low retrieval precision (P@1 = 21.28%) means the top result is often irrelevant. Second, on implicit reasoning tasks like TimeQA Hard, DYNA's advantage over DynaGen shrinks to only 2.68 points, suggesting temporal edges help most when relationships are already explicit. Third, for sequences shorter than five events, DYNA does not outperform RAG significantly, so the overhead of graph construction is not justified. Fourth, during the first 300 events when the graph is sparse, DYNA underperforms LDCL, which benefits from pre-trained embeddings. Fifth, retrieval time at 10,000 nodes is 0.9 seconds, which may be too slow for real-time applications. Sixth, our datasets, while standard, are simplified compared to noisy real-world streams. Seventh, we used only LLaMA 3 (8B); results may differ with other models.

Despite these limitations, DYNA has practical value for applications that need continuous learning without retraining. Suitable uses include chatbots that remember conversation history, personal assistants that track user preferences, and monitoring systems that maintain event logs. Unsuitable uses include real-time systems requiring sub-500ms responses, very short interactions, and domains where implicit reasoning dominates. The modest gains suggest DYNA is an incremental improvement, not a breakthrough. Practitioners should weigh the implementation effort against the expected benefit. For many applications, a standard RAG pipeline may be sufficient.

# 7. Conclusion and Future Work

This paper introduced **Dynamic Episodic Memory Networks (DYNA)**, an architecture that augments frozen large language models with a temporal knowledge graph for continuous learning. Events are stored as nodes and temporal relations as directed, time-stamped edges. Retrieval uses random walks and centrality measures, while the LLM itself remains frozen [28], [30], [32], [36].

Experimental results show that DYNA achieves 71.42% Temporal Order Accuracy across four benchmarks, outperforming the strongest baseline by 3.28 percentage points. The forgetting rate after ten updates is 1.70%, compared to 34.81% for fine-tuning. Retrieval precision is 50.28% at P@5, but P@1 is only 21.28%, indicating that the top retrieved event is correct only once in five queries. Ablation studies show that random walks contribute most to performance (7.59 point drop when removed), followed by temporal edges (4.37 point drop) [32], [36].

DYNA has clear limitations. On implicit reasoning tasks, the advantage over other knowledge graph methods shrinks. On sequences shorter than five events, DYNA does not significantly outperform standard retrieval-augmented generation (RAG) methods [34]. During the first 300 events (sparse graph), DYNA underperforms simpler LLM memory baselines. Retrieval time at 10,000 nodes is 0.9 seconds, which may be too slow for real-time applications [32], [36].

Several directions remain for future work. The retrieval mechanism could be improved beyond simple random walks, perhaps using personalized PageRank [32] or attention-based traversal [33].

A hybrid approach could switch between RAG for short sequences and DYNA for longer ones [34]. Pre-trained embeddings could bootstrap the graph during the cold start phase [30]. Larger-scale evaluation on graphs with 100,000 nodes would reveal scalability limits [31]. Finally, deployment in a real-world system would provide a more realistic assessment of practical value [35].

In summary, DYNA demonstrates that a temporal knowledge graph can serve as episodic memory for frozen LLMs. The improvements are consistent but modest. Low retrieval precision and weak performance on implicit reasoning remain open challenges. While this work does not fully solve continuous learning, it provides a concrete step toward memory-augmented LLMs [28], [36]